\documentclass[10pt,twocolumn,letterpaper]{article}

\usepackage{btas}
\usepackage{times}
\usepackage{epsfig}
\usepackage{graphicx}
\usepackage{amsmath}
\usepackage{amssymb}

\usepackage{verbatim}
\usepackage{booktabs} 
\usepackage{adjustbox}
\usepackage{colortbl} 
\usepackage{xcolor} 
\usepackage{xfrac}
\usepackage[]{algorithm2e}
\usepackage{subcaption}
\usepackage{graphicx,multirow}
\usepackage{array}

\usepackage{epsfig}
\usepackage{graphicx}
\usepackage{verbatim}
\usepackage{booktabs} 
\usepackage{colortbl}
\usepackage{mathtools}

\usepackage[norule,symbol,perpage]{footmisc}

\btasfinalcopy 

\ifbtasfinal\fi
 \makeatletter  
\def\ps@IEEEtitlepagestyle{  
\def\@oddfoot{\mycopyrightnotice}  
\def\@evenfoot{}  
}  
\def\mycopyrightnotice{  
{\hfill \footnotesize 978-1-7281-1522-1/19/\$31.00 \copyright 2019 IEEE\hfill}  
}  
\makeatother
\begin{document}

\title{Defending Against Adversarial Iris Examples Using Wavelet Decomposition}

\author{Sobhan Soleymani, Ali Dabouei, Jeremy Dawson, and Nasser M. Nasrabadi, {\it Fellow, IEEE}\\
West Virginia University\\
{\tt\small {\{ssoleyma, ad0046\}@mix.wvu.edu,}}
{\tt\small {\{jeremy.dawson, nasser.nasrabadi\}@mail.wvu.edu}}}

\maketitle
\thispagestyle{empty}

\begin{abstract}
Deep neural networks have presented impressive performance in biometric applications. However, their performance is highly at risk when facing carefully crafted input samples known as adversarial examples. In this paper, we present three defense strategies to detect adversarial iris examples. These defense strategies are based on wavelet domain denoising of the input examples by investigating each wavelet sub-band and removing the sub-bands that are most affected by the adversary. The first proposed defense strategy reconstructs multiple denoised versions of the input example through manipulating the mid- and high-frequency components of the wavelet domain representation of the input example and makes a decision upon the classification result of the majority of the denoised examples. The second and third proposed defense strategies aim to denoise each wavelet domain sub-band and determine the sub-bands that are most likely affected by the adversary using the reconstruction error computed for each sub-band. We test the performance of the proposed defense strategies against several attack scenarios and compare the results with five state of the art defense strategies.   
\end{abstract}
\section{Introduction}
Adversarial examples~\cite{kurakin2016adversarial} are data samples modified to fool machine learning classifiers. However, these modifications can be constructed to be perceptually indistinguishable for human observers~\cite{kurakin2016adversarial,carlini2017adversarial}. Therefore, the adversary can utilize these samples to conceal the identity of a subject. In addition, these examples can be deployed to fool a security system and provide access to unauthorized subjects by matching the adversarial examples to a specific or any other authorized subject. Adversarial examples are considered security threats since examples that are designed to be misclassified by one machine learning model can also be misclassified by other models~\cite{bruna2013intriguing}. Therefore, adversarial examples can be generated without the exact knowledge of the recognition framework. The countermeasures against adversarial attacks, which are donated as defense strategies, aim to either make the classifiers more robust to the adversarial attacks or detect the adversarial examples. 

There exist a very large variability of iris patterns among different persons due to the chaotic morphogenesis involved in the formation of the iris pattern~\cite{daugman2009iris}. Additionally, although externally visible, the iris is relatively stable over the time as a well-protected internal organ. As the result, among different biometric traits, iris images are the most reliable human identification trait. However, the adversarial examples to fool the iris recognition systems can be a major threat to the security systems, since many recognition and security applications widely rely on iris recognition systems. In this paper, we propose three defense strategies that are optimized to detect adversarial iris examples. To this end, we benefit from the fact that the low and low-mid frequencies wavelet components of an iris, consists of rich information for iris recognition and are robust to noises~\cite{kim2004iris}. In addition, the adversarial attacks manipulate the classifiers by adding high-frequency components to the input sample and using an $L_p$ norm constraint to control the amount of the distortion. 

In the first defense strategy, we focus on creating several denoised versions of the input image example, where each denoised version is constructed by setting some randomly chosen mid- and high-level wavelet sub-bands to zero, and classify each of them using the classifier used for the unperturbed benign examples. The second defense strategy considers removing the sub-bands that are most likely perturbed by the adversary. In this strategy, each sub-band of the input image example is denoised and the sub-bands that change the most after denoising are ignored in the reconstruction of the denoised version of the input image example. The third strategy removes the sub-bands that are most likely affected by the adversary and replaces the other sub-bands with their corresponding denoised version.

\begin{figure*}
\begin{center}
\includegraphics[width=1\linewidth]{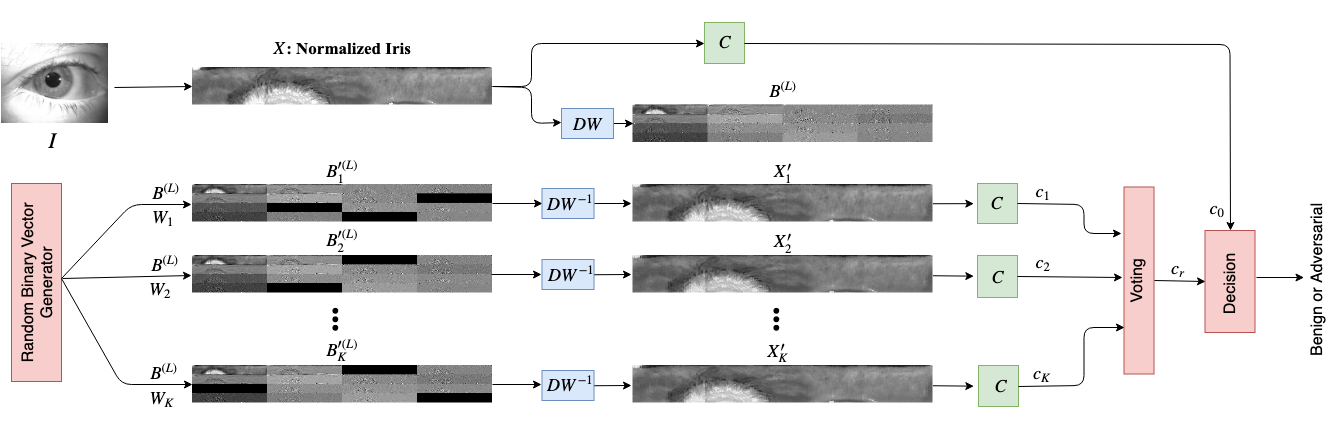}
\end{center}
\caption{In the first proposed defense strategy, the normalized iris image is considered as the input image example. $K$ denoised version of this example are generated through randomly selecting a maximum number of $N$ mid- and high-frequency wavelet sub-bands of the example and setting them to zero. Each denoised example is classified by classifier, $C$. The majority voting between classes assigned to the reconstructed examples by the classifier, $c_r$ is compared to the class assigned to the input image example, $c_0$, to decide whether or not the input image example is an adversarial example.}
\label{fig:fig1}
\end{figure*}

In our proposed frameworks we benefit from the fact that the wavelet decomposition divides the iris image into sub-bands that represent different vertical, horizontal, and diagonal frequency ranges. The low and low-mid frequency wavelet components of the iris are mainly used for iris recognition. These components, that are robust to noise, cannot be damaged drastically by the adversary to generate adversarial examples. 
In this paper, we make the following contributions: i) we introduce three defense strategies to recognize the adversarial iris images, ii) each of these strategies, that can be used as a preprocessing step in iris recognition frameworks, decompose the iris image into wavelet domain sub-bands, iii) the first strategy, randomly set some of the mid- and high-level sub-bands to zero, reconstruct several denoised versions of the input sample, and decide about the input example by classifying the denoised examples, iv) the second and third proposed strategies denoise each wavelet domain sub-band and determines the sub-bands affected by the adversary by investigating how much the sub-bands changes after denoising, and v) the proposed framework is robust to adversarial attacks and its performance is compared to several defense strategies.

\section{Related Works}
\subsection{Adversarial Attacks}
Deep learning models have outperformed the classical machine learning models in a variety of applications, such as biometrics~\cite{soleymani2018multi,soleymani2018generalized,soleymani2018prosodic}, security~\cite{talreja2017multibiometric,taherkhani2018deep}, and hashing~\cite{talreja2019learning,taherkhani2018facial,talreja2018using}. However, deep learning models are vulnerable to carefully crafted small perturbations in the input image. Although these small perturbations can change the predictions of the model, human observers cannot notice them. In other words, adversarial attacks~\cite{kurakin2016adversarial} aim to construct adversarial samples that can fool the classifier, while perceptually very similar to the benign samples~\cite{kurakin2016adversarial,carlini2017adversarial}. The adversary can utilize these samples to conceal the identity of a subject and acquire access to a biometric security system. One of the first adversarial attacks is generated by the authors in~\cite{szegedy2013intriguing} considering a L-BFGS method. Although this method is able to fool deep classifiers trained on different inputs, it is computationally expensive~\cite{dabouei2019fast}. 

The Fast Gradient Sign Method (FGSM)~\cite{goodfellow2014explaining} is proposed to compensate this shortcoming. The perturbation in FGSM is calculated based on the sign of the gradient of the classification loss with respect to the input sample. The authors in~\cite{rozsa2016adversarial} have increased the effectiveness of their attack by using the gradient value instead of the gradient sign. To reduce the computational cost of the attack, the authors in~\cite{papernot2016limitations} utilized a Jacobian matrix of the prediction of classes with respect to the input pixels. They have reduced the number of pixels that are required to be altered during the attack by calculating the saliency map of the input space. However, saliency-based methods are computationally expensive due to the greedy search for finding the most significant areas in the input sample. DeepFool~\cite{moosavi2016deepfool} finds $L_p$ minimal perturbations by iteratively translating input samples toward the closest decision boundary. 

\subsection{Adversarial Defense Strategies}
Defense strategies against adversarial examples can be categorized into two main categories~\cite{yuan2019adversarial}. Reactive strategies try to detect adversarial examples after deep neural   networks are built. On the other hand, proactive strategies make deep neural networks more robust before adversaries generate adversarial examples. Network distillation, adversarial training, and classifier robustifying are three major proactive defense methods. The authors in~\cite{papernot2016distillation} have considered distillation of the deep neural network~\cite{ba2014deep} to defend against adversarial examples. This approach is based on the fact that adversarial attacks on deep neural networks are successful because of the sensitivity of the deep networks. Therefore, reducing the sensitivity of the model using distillation, decreases the possibility of adversarial attacks. Authors in~\cite{goodfellow2014explaining} have included adversarial examples in their training phase. Although, this strategy increases the robustness of neural networks for one-step attacks but is not useful to avoid iterative attacks~\cite{kurakin2016adversarial}. Classifier robustifying aim to decrease the uncertainty from adversarial examples, employing different classifiers models~\cite{bradshaw2017adversarial}. The authors in~\cite{abbasi2017robustness} have observed that most adversarial examples are labeled as a small subset of incorrect classes. Therefore, to mitigate the misclassification effect of the adversarial examples, they have divided the classes into sub-classes and ensembled the result from sub-classes by majority voting.

Three major reactive strategies to prevent adversarial examples are adversarial detecting, input  reconstruction, and network verification. The authors in~\cite{gong2017adversarial} have considered training a binary classifier to detect the adversarial examples. The authors in~\cite{grosse2017statistical} have added an outliers class to their original classifier to detect the adversarial examples. The authors in~\cite{song2017pixeldefend} have observed that the distribution of the real data is different than the distribution of the adversarial data. 
The authors in~\cite{meng2017magnet} have observed that the adversarial examples have different low-ranked coefficients after principal component analysis compared to the benign examples. Input reconstruction aims to transform the adversarial examples to their corresponding benign examples, in order for the adversarial examples to be classified into their correct classes. The authors in~\cite{meng2017magnet} have trained a denoising auto-encoder network to transform adversarial examples to benign examples by removing the adversarial perturbations. Verifying the properties of deep neural networks is a reliable defense strategy, since it can detect the new unseen attacks. Network verification methods examine whether an input violates the properties of a neural network~\cite{katz2017reluplex}.
\begin{figure*}
\begin{center}
\includegraphics[width=1\linewidth]{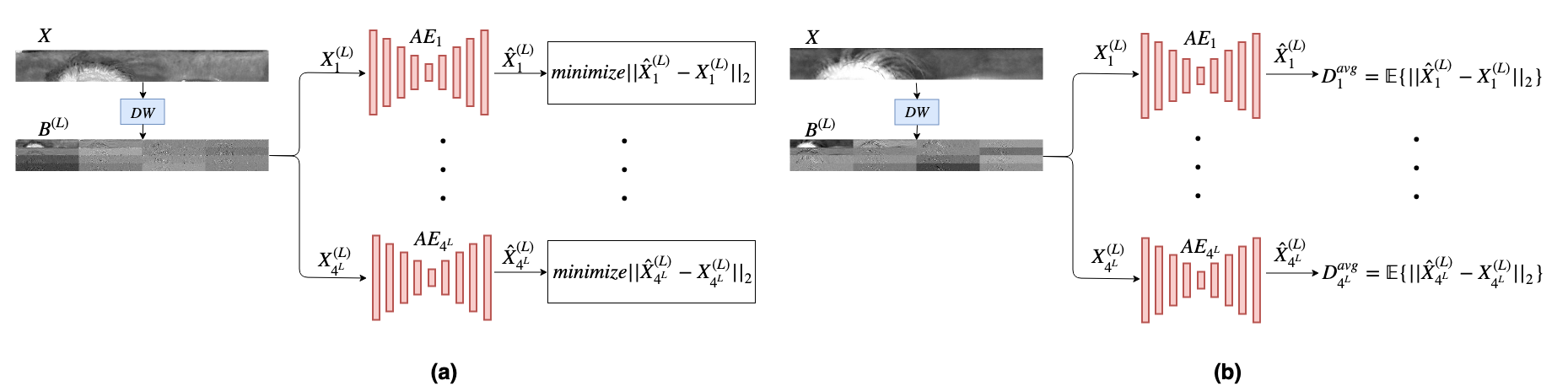}
\end{center}
\caption{Training the denoising auto-encoders for the second and third proposed defense strategies. (a) These auto-encoders are trained to reconstruct their corresponding wavelet sub-bands through the $L_2$ reconstruction loss. (b) The trained auto-encoders are utilized to compute the average reconstruction error for each sub-band on the validation set.}
\label{fig:fig21}
\end{figure*}
\section{Proposed Method}
In our proposed defense strategies, we do not alter the classifier, but aim to detect adversarial examples by removing perturbations in the input examples. In other words, the same classifier which is used for the unperturbed benign examples is considered to detect the adversarial examples by classifying the denoised version of the input image examples. To this aim, we decompose the input example into its corresponding uniform wavelet sub-bands using a uniform wavelet transform.  
Our first proposed defense strategy, randomly sets some of the mid-high and high frequency wavelet sub-bands to zero, and then reconstructs the iris image. The reconstructed iris image is classified using the classifier which is used for the benign iris images. This process is repeated several times using the same classifier. When the majority of the reconstructed iris images are classified as a different class compared to the class assigned to the iris image, the iris image is considered as an adversarial example. Since for each reconstructed iris image some of the wavelet sub-bands are randomly chosen and set to zero, the adversary cannot use this information to re-train their adversarial network.

The second and third proposed defense strategies, investigate each wavelet domain sub-band and determine the sub-bands that are most likely affected by the adversary. To this end, we train a denoising auto-encoder for each wavelet domain sub-band on the training set. Then, for each test set example we compute the sub-band specific reconstruction errors for all the sub-bands. The sub-bands that change the most after denoising are considered as the sub-bands that are most affected by the adversary. In the second defense strategy, these sub-bands are removed from the wavelet domain representation of the input example before reconstructing the denoised example. Similarly, the third strategy removes the sub-bands that are most likely affected by the adversary. However, it replaces the other sub-bands with their corresponding denoised version, prior to reconstruction of the input example. If the classification result of the denoised example is the same as the label assigned to the input example, the input example is considered as benign.

\subsection{Uniform Wavelet Decomposition}
In contrast with hierarchical wavelet decomposition which aims to decompose the low-pass sub-bands more finely as the number of decomposition levels increases, uniform wavelet decomposition consists of decomposing an input image uniformly into equal sub-bands. This uniform decomposition provides our proposed framework with more flexibility to choose mid- and high-frequency sub-bands. There are $4^L$ sub-bands in an $L$-level two-dimensional uniform wavelet decomposition. We denote these $4^L$ sub-bands by $B^{(L)}= \{X^{(L)}_1, X^{(L)}_2, \dotsc$, $X^{(L)}_{4^L}\}$, where $X^{(L)}_1$ represents the low-frequency component of the wavelet decomposition. Assume that $X$ is the input image example, and $g[n]$ and $h[n]$ are the low-pass and high-pass analysis wavelet filters, respectively. If $X_i^{(L-1)}$ is the $i^{th}$ sub-band in $(L-1)$-level uniform wavelet decomposition of the input image example, $X$, its four corresponding level $L$ sub-bands after another level of uniform wavelet decomposition are:
\begin{eqnarray}
\small
\begin{split}
X_{4i-3}^{(L)}[n_1,n_2]= \{X_i^{(L-1)}[n_1,n_2]*g[n_1]*g[n_2]\}2\downarrow_V 2\downarrow_H,\\
X_{4i-2}^{(L)}[n_1,n_2]= \{X_i^{(L-1)}[n_1,n_2]*h[n_1]*g[n_2]\}2\downarrow_V 2\downarrow_H,\\
X_{4i-1}^{(L)}[n_1,n_2]= \{X_i^{(L-1)}[n_1,n_2]*g[n_1]*h[n_2]\}2\downarrow_V 2\downarrow_H,\\
X_{4i}^{(L)}[n_1,n_2]= \{X_i^{(L-1)}[n_1,n_2]*h[n_1]*h[n_2]\}2\downarrow_V 2\downarrow_H,\\
\end{split}
\label{Eq:analysis}
\end{eqnarray}
where $n_1$ and $n_2$ represent horizontal and vertical indexes, respectively. $X^{(0)}_1=X$, and $2\downarrow_V$ and $2\downarrow_H$ represent down-sampling vertically and horizontally by a factor of $2$, respectively. 

After denoising the input image example, $X$, in the wavelet domain, the image $X'$ is reconstructed from sub-bands $B'^{(L)}= \{X'^{(L)}_1, X'^{(L)}_2, \dotsc, X'^{(L)}_{4^L}\}$, where $B'^{(L)}$ is the modified version of $B^{(L)}$. As described in Section~\ref{sectionStrategy}, this modification is done in order to denoise the input image. Assume that $g_1[n]$ and $h_1[n]$ are the low-pass and high-pass synthesis wavelet filters, respectively. Then, $X'^{(L-1)}_i$ which is the $i^{th}$ sub-band in the $(L-1)$-level uniform wavelet decomposition of image $X'$, is reconstructed from its four corresponding wavelet sub-bands in $L$-level wavelet decomposition:
\begin{equation}
\small
\begin{split}
X'^{(L-1)}_i[n_1,n_2]=\{X'^{(L)}_{4i-3}[n_1,n_2]*g_1[n_1]*g_1[n_2]\}2\uparrow_V 2\uparrow_H\\ 
+\{X'^{(L)}_{4i-2}[n_1,n_2]*h_1[n_1]*g_1[n_2]\}2\uparrow_V 2\uparrow_H\\
+\{X'^{(L)}_{4i-1}[n_1,n_2]*g_1[n_1]*h_1[n_2]\}2\uparrow_V 2\uparrow_H\\
+\{X'^{(L)}_{4i}[n_1,n_2]*h[n_1]*h[n_2]\}2\uparrow_V 2\uparrow_H,
\end{split}
\label{Eq:syntheisis}
\end{equation}
where $X'=X^{(0)}_1$, $2\uparrow_V$ and $2\uparrow_H$ represent up-sampling vertically and horizontally by a factor of $2$, respectively.
\begin{figure*}
\begin{center}
\includegraphics[width=1\linewidth]{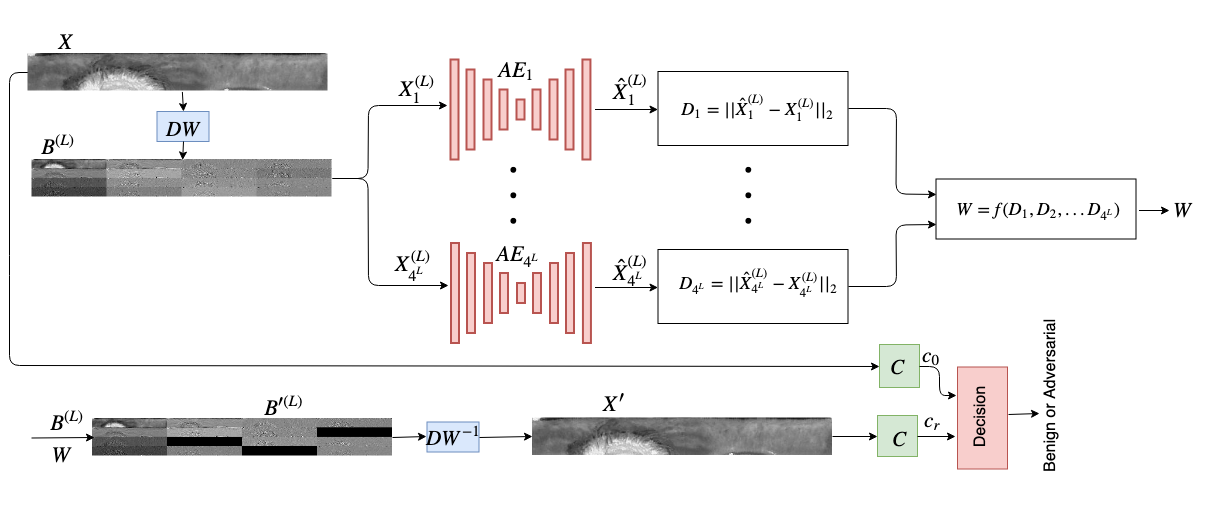}
\end{center}
\caption{Second and third proposed defense strategies: The trained auto-encoders are employed to denoise each wavelet sub-band of the input image example. The distance between each sub-band and its corresponding denoised sub-band, along with the average reconstruction error for each sub-band computed on the validation set, are considered to determine which sub-bands should be ignored in order to denoise the input image example. The second strategy, considers keeping the other sub-bands without any change, while the third strategy uses the denoised version of them in the reconstruction. The class assigned to the reconstructed example by the classifier, $c_r$, is compared to the class assigned to the input image example, $c_0$, to determine whether or not the input image example is an adversarial example.}
\label{fig:fig22}
\end{figure*}
\subsection{Defense Strategies}
\label{sectionStrategy}
The proposed defense strategies are effective for uniform wavelet decomposition and specifically iris images since each wavelet sub-band roughly includes a different range of vertical, horizontal, or diagonal frequencies. In the first proposed defense strategy, as presented in Figure~\ref{fig:fig1}, $K$ denoised versions of the input iris image, $X$, are reconstructed. We denote these reconstructed versions as $X'_i, i=1,2,...,K$. To construct each denoised iris image, $X'_i$, we decompose the input image example into its corresponding uniform wavelet sub-bands, $B^{(L)}= \{X^{(L)}_1, X^{(L)}_2, \dotsc$, $X^{(L)}_{4^L}\}$, using Equation~\ref{Eq:analysis}. A maximum number of $N$ high- and mid-level sub-bands are randomly selected and represented by binary vector $W_i\in \mathbb{R}^{4^L}$, where $0$s in this vector represent the sub-bands that should be set to zero and $1$s represent sub-bands that we keep. This vector along with the original wavelet sub-bands are used to construct sub-bands, $B'^{(L)}_i$, representing the denoised version of the original input iris image. Then, these sub-bands, $B'^{(L)}_i$, are utilized to reconstruct the denoised iris image, $X'_i$, using Equation~\ref{Eq:syntheisis}. Each reconstructed input image example, $X'_i$, is classified by the same classifier that is used for the unperturbed benign iris examples. If the majority of the $K$ reconstructed iris images are classified to the label assigned to the iris image, we consider the iris image as a benign example. Otherwise, the iris image is considered as an adversarial example. 

The second and third proposed defense strategies focus on removing the sub-bands that are perturbed the most by the adversarial attack. To this aim, as shown in Figure~\ref{fig:fig21}, we train a denoising auto-encoders for each sub-band using benign samples in the training set. Each auto-encoder, $AE_i$, which aims to reconstruct its corresponding sub-band, is trained on the $i^{th}$ sub-band of the benign examples, $X^{(L)}_i$. This auto-encoder reconstructs this sub-band as $\hat X^{(L)}_i$. We aim to minimize the difference between the input original sub-band and the reconstructed sub-band using the following loss function:
\begin{equation}
L^{rec}_i=||\hat X^{(L)}_i-X^{(L)}_i||_2.
\end{equation}
After training each auto-encoder, $AE_i$, we compute the average of the distance between sub-bands and the corresponding reconstructed sub-bands for the benign examples in the validation set as the sub-band specific average reconstruction error:
\begin{equation}
D^{avg}_i=\mathbb{E}\{||\hat X^{(L)}_i-X^{(L)}_i||_2\},
\end{equation}
where the expectation is calculated over the benign samples in the verification set. These distances, are employed to recognize the sub-bands that are removed or denoised when reconstructing the input image example. As presented in Figure~\ref{fig:fig22}, the trained auto-encoders are utilized to reconstruct the denoised version of the input example, $X$, denoted by $X'$. To this end, the input example is decomposed into its corresponding sub-bands. Each sub-band is fed into the corresponding denoising auto-encoder to be denoised. The reconstruction error for each sub-band is calculated as:
\begin{equation}
D_i=||\hat X^{(L)}_i-X^{(L)}_i||_2.
\end{equation}
These distances along with the average distances for the sub-bands on the training set, $D^{avg}_i$ are utilized to construct the binary vector $W$: 
\begin{equation}
\alpha_i=\frac{D_i}{D^{avg}_i},
\end{equation}
where the elements corresponding to the $N$ largest $\alpha$ values are set to zero for generating the binary vector $W$. This vector along with $B^{(L)}$ is utilized to construct $B'^{(L)}$ and consequently $X'$. If this image example is classified similar to the input image example, $X$, the iris image is considered as a benign example. Otherwise, it is considered as an adversarial examples. The third defense strategy replaces the sub-bands corresponding to the other $4^L-N$ values with their denoised version before reconstructing the input image.
\section{Experimental Setup}
In this section, we describe the attack scenarios, the dataset, and the optimization methods. We conclude the section with reporting the results for the proposed frameworks and comparing their performance with the state-of-the-art frameworks. 
\begin{figure*}
\begin{center}
\includegraphics[width=1\linewidth]{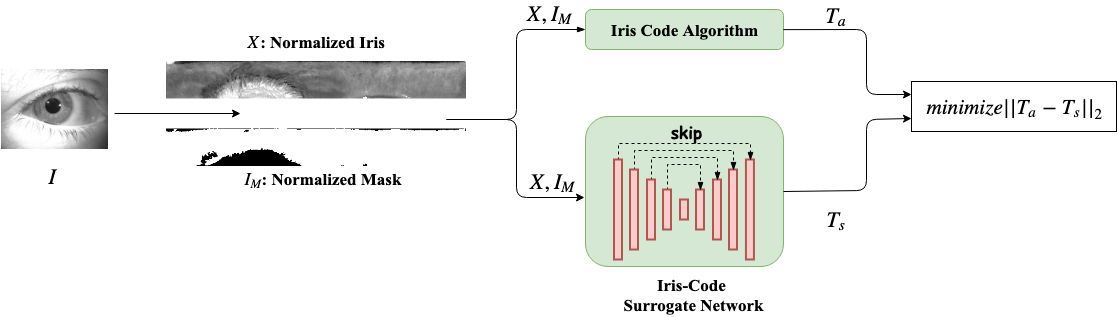}
\end{center}
\caption{Iris code generation: The normalized iris image and the normalized mask are concatenated in depth as the input to the iris code surrogate deep network.,The output is forced to mimic the iris code by minimizing the reconstruction loss. The trained surrogate network is utilized to generate adversarial examples.}
\label{fig:surrogate}
\end{figure*}
\subsection{Attack Scenarios}
The majority of the iris identification frameworks consider constructing iris-codes from the iris image~\cite{daugman2009iris,szewczyk2012reliable,krichen2008osiris,masek2003matlab}. The iris-code is generally constructed through segmentation, mask generation, normalization, and binerization. We consider the normalized iris images as the image examples. In the conventional iris identifications frameworks, the normalized iris image is converted to an iris template through multiple levels of 2-D Gabor or wavelets filters. In these frameworks, during the authentication or recognition algorithm, iris-codes which are the constructed by binarizing the iris templates are compared using bit-based metrics such as Hamming distance. 

However, conventional filter bank-based iris-code generation frameworks cannot be employed in our experimental setup to generate adversarial examples since generating adversarial examples requires back-propagation of the adversarial loss. Therefore, to compensate for this shortcoming, we train an auto-encoder surrogate network to mimic the conventional iris code generation procedure. As presented in Figure~\ref{fig:surrogate}, the normalized iris image and the normalized iris mask image, which are concatenated in depth, are fed to this surrogate network to generate the iris code by minimizing the reconstruction loss. In this figure the template generated by the conventional iris verification algorithm is denoted by $T_a$ and the iris template generated by the surrogate network is denoted by $T_s$. The reconstruction loss is defined as the $L_2$ distance between these two templates~\cite{soleymani2019adversarial}. The architecture for this surrogate network is presented in Table~\ref{table:architecture_table}.

The trained surrogate network, which follows the U-net architecture~\cite{ronneberger2015u}, is then utilized to generate the adversarial examples from the test set using the fast gradient sign method algorithm (FGSM)~\cite{goodfellow2014explaining}, iterative gradient sign method algorithm (iGSM)~\cite{kurakin2016adversarial}, and Deepfool~\cite{moosavi2016deepfool}. The step-size for iGSM algorithm is set $0.005$. Following the framework in~\cite{rathgeb2017feasibility}, the normalized iris images for which the Hamming distance between their iris-code and their corresponding benign iris-code is more than $32\%$ are considered as adversarial examples. This assumption results in False Match Rate of about $0.0001\%$~\cite{rathgeb2017feasibility}. We consider this criteria as the termination criteria for the adversarial attacks.  

\begin{table}[t]
\caption[Table caption text]{Iris-code deep surrogate network: In this architecture, which follows a U-Net architecture, the first five layers represent the encoding sub-network, while the next five layers are the decoding layers. Conv and deconv represent convolutional and deconvolutional layers, respectively.}  
\begin{center}
\addtolength{\tabcolsep}{-5pt}
\begin{tabular}{l@{\hskip .05in}c@{\hskip .05in}c@{\hskip .05in}c}

\hline
layer&kernel&input&output\\
\hline
conv1&   $4{\times} 4 {\times} 64$ & $64{\times} 512 {\times} 2$& $32{\times} 256{\times} 64$\\
\rowcolor{black!10}conv2&   $4{\times} 4 {\times} 128$& $32{\times} 256 {\times} 64$&$16{\times} 128 {\times} 128$\\
conv3&   $4{\times} 4 {\times} 256$& $16{\times} 128 {\times} 128$& $8{\times} 64 {\times} 256$\\
\rowcolor{black!10}conv4&   $4{\times} 4 {\times} 512$& $8{\times}  64  {\times} 256$& $4{\times} 32 {\times} 512$\\
conv5&   $4{\times} 4 {\times} 512$& $4{\times}  32  {\times} 512$& $2{\times} 16 {\times} 512$\\
\hline
\rowcolor{black!10}deconv4& $4{\times} 4 {\times} 512$& $2{\times}  16  {\times} 512$& $4{\times} 32 {\times} 512$\\
deconv3& $4{\times} 4 {\times} 256$& $4{\times}  32 {\times} (512+512)$& $8{\times} 64 {\times} 256$\\
\rowcolor{black!10}deconv2& $4{\times} 4 {\times} 128$& $8{\times} 64 {\times} (256+256)$&$16{\times} 128 {\times} 128$\\
deconv1& $4{\times} 4 {\times} 64$ & $16{\times} 128 {\times} (128+128)$& $32{\times} 256{\times} 64$\\
\rowcolor{black!10}deconv0& $4{\times} 4 {\times} 6$ & $32{\times} 256 {\times} (64+64)$& $64{\times} 512{\times} 6$\\
\bottomrule
\end{tabular}
\end{center}
\label{table:architecture_table}
\end{table}

\begin{table}[t]
\caption[Table caption text]{Denoising auto-encoder network architecture: The first three layers represent the encoding sub-network, while the next three layers are the decoding layers. Conv and deconv represent convolutional and deconvolutional layers, respectively.}  
\begin{center}
\addtolength{\tabcolsep}{-5pt}
\begin{tabular}{l@{\hskip .05in}c@{\hskip .05in}c@{\hskip .05in}c}

\hline
layer&kernel&input&output\\
\hline
conv1&   $4{\times} 4 {\times} 64$ & $16{\times} 128 {\times} 1$& $8{\times} 64{\times} 64$\\
\rowcolor{black!10}conv2&   $4{\times} 4 {\times} 128$& $8{\times} 64 {\times} 64$&$4{\times} 32 {\times} 128$\\
conv3&   $4{\times} 4 {\times} 256$& $4{\times} 32 {\times} 128$& $2{\times} 16 {\times} 256$\\
\hline
\rowcolor{black!10}deconv2& $4{\times} 4 {\times} 128$& $2{\times}  16  {\times} 256$& $4{\times} 32 {\times} 128$\\
deconv1& $4{\times} 4 {\times} 64$& $4{\times}  32 {\times} 128$& $8{\times} 64 {\times} 64$\\
\rowcolor{black!10}deconv0& $4{\times} 4 {\times} 1$& $8{\times} 64 {\times} 64$&$16{\times} 128 {\times} 1$\\

\bottomrule
\end{tabular}
\end{center}
\label{table:architecture_DAE}
\end{table}

\subsection{Dataset and Optimization}
In the experimental setup, the OSIRIS algorithm~\cite{krichen2008osiris} is considered to generate the normalized iris images, normalized iris masks, and iris-codes. The normalized iris images and normalized iris masks are of size $64\times 512$. OSIRIS algorithm considers a filter bank of six Gabor filters to generate the binary iris-codes of size $384 \times 512$. Our classifier algorithm utilizes the iris-codes generated by OSIRIS. In our framework, ADAM solver for stochastic optimization~\cite{kingma2014adam} is used to train the surrogate network and denoising auto-encoders. All the optimizations are conducted using learning rate of $10^{-4}$. For the surrogate network and denoising auto-encoders, the encoding and decoding sub-networks are trained with $2\times 2$ and $1\times 1$ stride sizes, respectively. For the surrogate network, the the encoding layers are concatenated in depth with the corresponding layers in the decoding sub-network. Separable kernels~\cite{szegedy2015going} are considered for all the layers. The networks are trained using mini-batch of size $64$. Batch normalization is applied on the outputs of all the layers. A ReLU activation function is utilized for all the layers except the deconv0 layer in the surrogate network, where $tanh$ is considered. For the surrogate network, the $64{\times} 512{\times} 6$ output is reshaped to $384 \times 512$ to be compatible to the iris-code.

Two dataset are considered in our experimental setup. The iris-code surrogate network and denoising auto-encoders, which are utilized in the second and third defense strategies, are trained on $8,000$ normalized iris images from the BioCop dataset~\cite{BIIC}. Denoising auto-encoders are utilized using the verification set to compute average reconstruction distance for the sub-bands. The details of the denoising auto-encoder architecture are presented in Table~\ref{table:architecture_DAE}. The verification set consists of $2,000$ iris images from the BioCop dataset. The test set is constructed using $3,040$ iris images from $231$ subjects in the BIOMDATA dataset~\cite{crihalmeanu2007protocol}. One iris image from each subject is considered as the gallery and the other samples are used as the probe. The Hamming distance between the iris-codes corresponding to the iris images in the gallery and prob are considered as our classification criteria. To make the experimental setup unbiased, we consider the test set to include $50\%$ benign and $50\%$ adversarial examples.  
\subsection{Experimental Results}
We compare our proposed defense strategies with five state-of-the-art frameworks. We consider the success rate of recognizing the adversarial and benign examples as the evaluation metric for defense strategies. Three adversarial training frameworks~\cite{goodfellow2014explaining,tramer2017ensemble,madry2017towards} and two denoising frameworks~\cite{meng2017magnet,shaham2018defending} are considered as the baselines for the proposed strategies. MagNet~\cite{meng2017magnet} considers using a denoising auto-encoder and~\cite{shaham2018defending} denoises the input image example using JPEG compression. For all our experiments, we consider Haar wavelet and two levels of uniform wavelet decomposition which results in $16$ sub-bands. 

In our first experimental setup, we focus on the first proposed defense strategy. In this setup, we investigate how the maximum number of sub-bands that are set to zero, $N$, and the number of examples reconstructed, $K$, affect the success rate. For this strategy, we consider that twelve mid- and high-frequency sub-bands can be set to zero. As presented in Table~\ref{table:strategy1}, the performance of the proposed strategy improves when the number of reconstructed examples increases. On the other hand, when increasing the maximum number of sub-bands that are set to zero, the performance of this strategy drops after $K$ equals to five sub-bands. 

The second experimental setup investigates the performance of the second and third defense strategies when the number of sub-bands that are set to zero varies. As presented in Table~\ref{table:strategy2}, similar to the first strategy, the performance first increases and then drops. As presented in these two tables, both these strategies outperforms the first strategy. This is due to the fact that randomly selecting and forcing some of the sub-bands to zero, may destroy some useful information as well as the sub-bands which are not affected by the adversary. This may increase the false rejection rate. However, when we select the bands affected by the adversary based on the reconstruction error, we aim to keep the information not destroyed by the adversary. Therefore, the overall performance of the defense strategy improves. In this table, we also present the average performance of the classifier for no attack scenario. As expected, this performance decreases when the number of sub-bands set to zero is increased.  

Table~\ref{table:comparsion} presents the performance of five state of the art defense frameworks on the test set. As presented in this table, the adversarial training algorithms are outperformed by the denoising frameworks. In addition, our proposed defense strategies outperform both denoising algorithms. This is due to the fact that instead of using a single auto-encoder~\cite{meng2017magnet} or denoising the input image example through compression~\cite{shaham2018defending}, we aim to figure out which sub-bands are most likely affected by the adversary. By defining the sub-band specific distances, we customize the algorithm for each input image example. In other words, the proposed defense strategies do not denoise the input example blindly, but specifically uses the sub-bands affected by the adversary to denoise the input example. 

\begin{table}[t]
\caption[Table caption text]{The performance of the first defense strategy for the FGSM adversarial attack, when the maximum number of sub-bands that are set to zero, $N$, and the number of examples reconstructed, $K$, are varied.}  
\begin{center}
\addtolength{\tabcolsep}{-5pt}
\begin{tabular}{l@{\hskip .05in}|c@{\hskip .05in}c@{\hskip .05in}c@{\hskip .05in}c@{\hskip .05in}c@{\hskip .05in}c@{\hskip .05in}c}
\hline
$K$, $N$ &1&2&3&4&5&6&7\\
\hline
\rowcolor{black!10} 3& 12.50& 20.87 &22.21 &26.67  &32.38&28.21&26.31\\
				    5& 22.13& 24.54 &31.92 &40.48  &45.58&40.67&35.56\\
\rowcolor{black!10} 7& 24.79& 26.91 &35.72 &43.21  &56.37&51.90&44.75\\
				   10& 26.87& 28.73 &38.64 &54.91  &62.75&59.12&49.43\\
\rowcolor{black!10}15& 28.31& 30.84 &42.83 &64.43  &64.81&64.37&54.57\\
				   20& 31.10& 35.51 &62.75 &72.97  &75.01&65.89&65.23\\
\rowcolor{black!10}30& 32.14& 46.87 &63.81 &72.41  &76.08&66.08&65.46\\
\bottomrule
\end{tabular}
\end{center}
\label{table:strategy1}
\end{table}

\begin{table}[t]
\caption[Table caption text]{The performance of the second and third defense strategies for three adversarial attacks when the number of sub-bands that are set to zero, $N$, is varied. }  
\begin{center}
\small
\addtolength{\tabcolsep}{-5pt}
\begin{tabular}{l@{\hskip .05in}|c@{\hskip .05in}|c@{\hskip .05in}c@{\hskip .05in}c@{\hskip .05in}|c@{\hskip .05in}c@{\hskip .05in}c}
\hline
\multicolumn{2}{c|}{}&\multicolumn{3}{c|}{Ours$\#2$}&\multicolumn{3}{c}{Ours$\#3$}\\
\hline
$N$&No Attack&FGSM&iGSM&Deepfool&FGSM&iGSM&Deepfool\\
\hline
\rowcolor{black!10}1&99.10& 12.57&  8.31 &18.35&    12.91& 10.42 &18.54\\
				   2&98.97& 15.19& 12.67 &25.48&    15.24& 14.86 &25.92\\
\rowcolor{black!10}3&98.86& 27.38& 20.72 &37.53&    28.51& 23.12 &37.73\\
				   4&98.54& 38.52& 32.15 &63.74&    39.54& 35.28 &64.11\\
\rowcolor{black!10}5&98.21& 61.74& 55.43 &84.36&    62.12& 57.74 &84.36\\
				   6&98.07& 81.23& 75.59 &78.21&    81.65& 77.59 &78.25\\
\rowcolor{black!10}7&97.87& 78.18& 71.81 &70.42&    78.53& 73.81 &70.51\\
\bottomrule
\end{tabular}
\end{center}
\label{table:strategy2}
\end{table}

\begin{table}[t]
\caption[Table caption text]{The performance of proposed defense strategies compared to five state of the art algorithms. The first three algorithms are adversarial training frameworks, while the other two algorithms are denoising frameworks.}  
\begin{center}
\addtolength{\tabcolsep}{-5pt}
\begin{tabular}{l@{\hskip .05in}c@{\hskip .05in}c@{\hskip .05in}c}

\hline
 &FGSM&iGSM&Deepfool\\
\hline
\rowcolor{black!10}\cite{goodfellow2014explaining}&38.98&33.78&45.47\\
\cite{tramer2017ensemble}&37.87&34.97&44.41\\
\rowcolor{black!10}\cite{madry2017towards}&39.51&42.18&56.78\\
\cite{shaham2018defending}&45.15 &47.89&51.24\\
\rowcolor{black!10}\cite{meng2017magnet}& 57.08& 53.26&60.54\\
\hline
Ours$\#1$& 76.08& 71.26&79.54\\
\rowcolor{black!10}Ours$\#2$& 81.23& 75.59 &84.21\\
Ours$\#3$&81.65& 77.59 &84.36\\
\bottomrule
\end{tabular}
\end{center}
\label{table:comparsion}
\end{table}

\section{Conclusions}
We presented three defense strategies to detect the adversarial iris examples. These strategies investigate each wavelet sub-band in order to denoise the input examples. Through defining these defense strategies, we remove the sub-bands that are the most affected by the adversary. The first proposed defense strategy reconstructs multiple denoised versions of the input example. Then, this strategy decides about the input example through the classification of the denoised examples. The second proposed defense strategy denoises each wavelet domain sub-band. The sub-bands that are most likely affected by the adversary are determined by the $L_2$ distance between the wavelet domain sub-bands and their reconstructed version. Finally, the third proposed strategy, focuses on removing the sub-bands that are most effected by the adversary, while reconstructing the other sub-bands. We investigated the performance of the proposed defense strategies using three attack scenarios and compare the results with five state of the art defense strategies. These five strategies include the adversarial training and denoising frameworks. Our third proposed defense strategy, which aims to customize the sub-bands removed for each input example, outperforms the other two proposed defense strategies.     
\begin{center}
ACKNOWLEDGEMENT
\end{center}
This work is based upon a work supported by the Center for Identification Technology Research and the National Science Foundation under Grant $\#1650474$.
{\small
\bibliographystyle{ieee}
\bibliography{bib}

\begin{thebibliography}{10}\itemsep=-1pt

\bibitem{BIIC}
Biocop database, http://biic.wvu.edu/.

\bibitem{abbasi2017robustness}
M.~Abbasi and C.~Gagn{\'e}.
\newblock Robustness to adversarial examples through an ensemble of
  specialists.
\newblock {\em arXiv preprint arXiv:1702.06856}, 2017.

\bibitem{ba2014deep}
J.~Ba and R.~Caruana.
\newblock Do deep nets really need to be deep?
\newblock In {\em Advances in neural information processing systems}, pages
  2654--2662, 2014.

\bibitem{bradshaw2017adversarial}
J.~Bradshaw, A.~G. d.~G. Matthews, and Z.~Ghahramani.
\newblock Adversarial examples, uncertainty, and transfer testing robustness in
  gaussian process hybrid deep networks.
\newblock {\em arXiv preprint arXiv:1707.02476}, 2017.

\bibitem{bruna2013intriguing}
J.~Bruna, C.~Szegedy, I.~Sutskever, I.~Goodfellow, W.~Zaremba, R.~Fergus, and
  D.~Erhan.
\newblock Intriguing properties of neural networks.
\newblock {\em International Conference on Learning Representations}, 2014.

\bibitem{carlini2017adversarial}
N.~Carlini and D.~Wagner.
\newblock Adversarial examples are not easily detected: Bypassing ten detection
  methods.
\newblock In {\em Proceedings of the 10th ACM Workshop on Artificial
  Intelligence and Security}, pages 3--14. ACM, 2017.

\bibitem{crihalmeanu2007protocol}
S.~Crihalmeanu, A.~Ross, S.~Schuckers, and L.~Hornak.
\newblock A protocol for multibiometric data acquisition, storage and
  dissemination.
\newblock {\em Technical Report, WVU, Lane Department of Computer Science and
  Electrical Engineering}, 2007.

\bibitem{dabouei2019fast}
A.~Dabouei, S.~Soleymani, J.~Dawson, and N.~Nasrabadi.
\newblock Fast geometrically-perturbed adversarial faces.
\newblock In {\em 2019 IEEE Winter Conference on Applications of Computer
  Vision (WACV)}, pages 1979--1988, 2019.

\bibitem{daugman2009iris}
J.~Daugman.
\newblock How iris recognition works.
\newblock In {\em The essential guide to image processing}, pages 715--739.
  2009.

\bibitem{gong2017adversarial}
Z.~Gong, W.~Wang, and W.-S. Ku.
\newblock Adversarial and clean data are not twins.
\newblock {\em arXiv preprint arXiv:1704.04960}, 2017.

\bibitem{goodfellow2014explaining}
I.~J. Goodfellow, J.~Shlens, and C.~Szegedy.
\newblock Explaining and harnessing adversarial examples.
\newblock {\em arXiv preprint arXiv:1412.6572}, 2014.

\bibitem{grosse2017statistical}
K.~Grosse, P.~Manoharan, N.~Papernot, M.~Backes, and P.~McDaniel.
\newblock On the (statistical) detection of adversarial examples.
\newblock {\em arXiv preprint arXiv:1702.06280}, 2017.

\bibitem{katz2017reluplex}
G.~Katz, C.~Barrett, D.~L. Dill, K.~Julian, and M.~J. Kochenderfer.
\newblock Reluplex: An efficient {SMT} solver for verifying deep neural
  networks.
\newblock In {\em International Conference on Computer Aided Verification},
  pages 97--117, 2017.

\bibitem{kim2004iris}
J.~Kim, S.~Cho, J.~Choi, and R.~J. Marks.
\newblock Iris recognition using wavelet features.
\newblock {\em Journal of VLSI signal processing systems for signal, image and
  video technology}, 38(2):147--156, 2004.

\bibitem{kingma2014adam}
D.~P. Kingma and J.~Ba.
\newblock Adam: A method for stochastic optimization.
\newblock {\em arXiv preprint arXiv:1412.6980}, 2014.

\bibitem{krichen2008osiris}
E.~Krichen, A.~Mellakh, S.~Salicetti, and B.~Dorizzi.
\newblock Osiris (open source for iris) reference system.
\newblock {\em BioSecure Project}, 2008.

\bibitem{kurakin2016adversarial}
A.~Kurakin, I.~Goodfellow, and S.~Bengio.
\newblock Adversarial examples in the physical world.
\newblock {\em International Conference on Learning Representations-Workshop},
  2017.

\bibitem{madry2017towards}
A.~Madry, A.~Makelov, L.~Schmidt, D.~Tsipras, and A.~Vladu.
\newblock Towards deep learning models resistant to adversarial attacks.
\newblock {\em arXiv preprint arXiv:1706.06083}, 2017.

\bibitem{masek2003matlab}
L.~Masek and P.~Kovesi.
\newblock Matlab source code for a biometric identification system based on
  iris patterns.
\newblock 2003.

\bibitem{meng2017magnet}
D.~Meng and H.~Chen.
\newblock Magnet: a two-pronged defense against adversarial examples.
\newblock In {\em Proceedings of the 2017 ACM SIGSAC Conference on Computer and
  Communications Security}, pages 135--147. ACM, 2017.

\bibitem{moosavi2016deepfool}
S.-M. Moosavi-Dezfooli, A.~Fawzi, and P.~Frossard.
\newblock Deepfool: a simple and accurate method to fool deep neural networks.
\newblock In {\em Proceedings of the IEEE conference on computer vision and
  pattern recognition}, pages 2574--2582, 2016.

\bibitem{papernot2016limitations}
N.~Papernot, P.~McDaniel, S.~Jha, M.~Fredrikson, Z.~B. Celik, and A.~Swami.
\newblock The limitations of deep learning in adversarial settings.
\newblock In {\em Security and Privacy (EuroS\&P), 2016 IEEE European Symposium
  on}, pages 372--387, 2016.

\bibitem{papernot2016distillation}
N.~Papernot, P.~McDaniel, X.~Wu, S.~Jha, and A.~Swami.
\newblock Distillation as a defense to adversarial perturbations against deep
  neural networks.
\newblock In {\em 2016 IEEE Symposium on Security and Privacy (SP)}, pages
  582--597, 2016.

\bibitem{rathgeb2017feasibility}
C.~Rathgeb and C.~Busch.
\newblock On the feasibility of creating morphed iris-codes.
\newblock In {\em 2017 IEEE International Joint Conference on Biometrics
  (IJCB)}, pages 152--157, 2017.

\bibitem{ronneberger2015u}
O.~Ronneberger, P.~Fischer, and T.~Brox.
\newblock U-net: Convolutional networks for biomedical image segmentation.
\newblock In {\em International Conference on Medical image computing and
  computer-assisted intervention}, pages 234--241, 2015.

\bibitem{rozsa2016adversarial}
A.~Rozsa, E.~M. Rudd, and T.~E. Boult.
\newblock Adversarial diversity and hard positive generation.
\newblock In {\em Proceedings of the IEEE Conference on Computer Vision and
  Pattern Recognition Workshops}, pages 25--32, 2016.

\bibitem{shaham2018defending}
U.~Shaham, J.~Garritano, Y.~Yamada, E.~Weinberger, A.~Cloninger, X.~Cheng,
  K.~Stanton, and Y.~Kluger.
\newblock Defending against adversarial images using basis functions
  transformations.
\newblock {\em arXiv preprint arXiv:1803.10840}, 2018.

\bibitem{soleymani2019adversarial}
S.~Soleymani, A.~Dabouei, J.~Dawson, and N.~M. Nasrabadi.
\newblock Adversarial examples to fool iris recognition systems.
\newblock {\em arXiv preprint arXiv:1906.09300}, 2019.

\bibitem{soleymani2018prosodic}
S.~Soleymani, A.~Dabouei, S.~M. Iranmanesh, H.~Kazemi, J.~Dawson, and N.~M.
  Nasrabadi.
\newblock Prosodic-enhanced siamese convolutional neural networks for
  cross-device text-independent speaker verification.
\newblock {\em arXiv preprint arXiv:1808.01026}, 2018.

\bibitem{soleymani2018multi}
S.~Soleymani, A.~Dabouei, H.~Kazemi, J.~Dawson, and N.~M. Nasrabadi.
\newblock Multi-level feature abstraction from convolutional neural networks
  for multimodal biometric identification.
\newblock In {\em 24th International Conference on Pattern Recognition (ICPR)},
  pages 3469--3476, 2018.

\bibitem{soleymani2018generalized}
S.~Soleymani, A.~Torfi, J.~Dawson, and N.~M. Nasrabadi.
\newblock Generalized bilinear deep convolutional neural networks for
  multimodal biometric identification.
\newblock In {\em 25th IEEE International Conference on Image Processing},
  pages 763--767, 2018.

\bibitem{song2017pixeldefend}
Y.~Song, T.~Kim, S.~Nowozin, S.~Ermon, and N.~Kushman.
\newblock Pixeldefend: Leveraging generative models to understand and defend
  against adversarial examples.
\newblock {\em arXiv preprint arXiv:1710.10766}, 2017.

\bibitem{szegedy2015going}
C.~Szegedy, W.~Liu, Y.~Jia, P.~Sermanet, S.~Reed, D.~Anguelov, D.~Erhan,
  V.~Vanhoucke, and A.~Rabinovich.
\newblock Going deeper with convolutions.
\newblock In {\em Proceedings of the IEEE conference on computer vision and
  pattern recognition}, pages 1--9, 2015.

\bibitem{szegedy2013intriguing}
C.~Szegedy, W.~Zaremba, I.~Sutskever, J.~Bruna, D.~Erhan, I.~Goodfellow, and
  R.~Fergus.
\newblock Intriguing properties of neural networks.
\newblock {\em arXiv preprint}, 2013.

\bibitem{szewczyk2012reliable}
R.~Szewczyk, K.~Grabowski, M.~Napieralska, W.~Sankowski, M.~Zubert, and
  A.~Napieralski.
\newblock A reliable iris recognition algorithm based on reverse biorthogonal
  wavelet transform.
\newblock {\em Pattern Recognition Letters}, 33(8):1019--1026, 2012.

\bibitem{taherkhani2018deep}
F.~Taherkhani, N.~M. Nasrabadi, and J.~Dawson.
\newblock A deep face identification network enhanced by facial attributes
  prediction.
\newblock In {\em Proceedings of the IEEE Conference on Computer Vision and
  Pattern Recognition Workshops}, pages 553--560, 2018.

\bibitem{taherkhani2018facial}
F.~Taherkhani, V.~Talreja, H.~Kazemi, and N.~Nasrabadi.
\newblock Facial attribute guided deep cross-modal hashing for face image
  retrieval.
\newblock In {\em 2018 International Conference of the Biometrics Special
  Interest Group (BIOSIG)}, pages 1--6, 2018.

\bibitem{talreja2019learning}
V.~Talreja, S.~Soleymani, M.~C. Valenti, and N.~M. Nasrabadi.
\newblock Learning to authenticate with deep multibiometric hashing and neural
  network decoding.
\newblock {\em arXiv preprint arXiv:1902.04149}, 2019.

\bibitem{talreja2018using}
V.~Talreja, F.~Taherkhani, M.~C. Valenti, and N.~M. Nasrabadi.
\newblock Using deep cross modal hashing and error correcting codes for
  improving the efficiency of attribute guided facial image retrieval.
\newblock In {\em 2018 IEEE Global Conference on Signal and Information
  Processing (GlobalSIP)}, pages 564--568, 2018.

\bibitem{talreja2017multibiometric}
V.~Talreja, M.~C. Valenti, and N.~M. Nasrabadi.
\newblock Multibiometric secure system based on deep learning.
\newblock In {\em 2017 IEEE Global conference on signal and information
  processing (globalSIP)}, pages 298--302, 2017.

\bibitem{tramer2017ensemble}
F.~Tram{\`e}r, A.~Kurakin, N.~Papernot, I.~Goodfellow, D.~Boneh, and
  P.~McDaniel.
\newblock Ensemble adversarial training: Attacks and defenses.
\newblock {\em arXiv preprint arXiv:1705.07204}, 2017.

\bibitem{yuan2019adversarial}
X.~Yuan, P.~He, Q.~Zhu, and X.~Li.
\newblock Adversarial examples: Attacks and defenses for deep learning.
\newblock {\em IEEE transactions on neural networks and learning systems},
  2019.

\end{thebibliography}
}

\end{document}